\documentclass{article}

\usepackage{PRIMEarxiv}
\usepackage[utf8]{inputenc} 
\usepackage[T1]{fontenc}    
\usepackage{hyperref}       
\usepackage{url}            
\usepackage{booktabs}       
\usepackage{amsfonts}       
\usepackage{amsmath}
\usepackage{nicefrac}       
\usepackage{microtype}      
\usepackage{lipsum}
\usepackage{algorithm}
\usepackage[noend]{algpseudocode}
\usepackage{graphicx}
\graphicspath{{media/}}     
\usepackage{amsmath}  
\usepackage{pgfplots}
\pgfplotsset{compat=1.16}
\usepackage{wrapfig}

\makeatletter
\def\BState{\State\hskip-\ALG@thistlm}
\makeatother

\title{Cross-Axis Transformer with 3D Rotary Positional Embedding
\thanks{\textit{\underline{Citation}}: 
\textbf{Authors. Title. Pages.... DOI:000000/11111.}} 
}

\author{
  Lily Erickson \\
  EmerGen LLC \\
  Minneapolis\\
  \texttt{lilyerickson@emergenlabs.com} \\
}
\begin{document}
\maketitle

\begin{abstract}
Despite lagging behind their modal cousins in many respects, Vision Transformers have provided an interesting opportunity to bridge the gap between sequence modeling, and image modeling. Up until now, however, vision transformers have largely been held back, due to both computational inefficiency, and lack of proper handling of spatial dimensions. In this paper, we introduce the Cross-Axis Transformer. CAT is a model inspired both by Axial Transformers, and Microsoft's recent Retentive Network, that drastically reduces the required number of floating point operations required to process an image, while simultaneously converging faster and more accurately than the Vision Transformers it replaces.
\end{abstract}


\section{Introduction}
There have been many attempts at processing visual data using neural networks. Convolution Neural Networks use learned kernels to allow local patches of data to share information with each other. ResNets perform a similar role, but along a different axis, attempting to carry a residual version of their state forward.

Recently, transformers have broken onto the scene, which instead of caring strictly about their neighboring features, instead attempt to create a probability distribution over the entire feature space, using a process known as attention.

\begin{equation}
Attention(Q, K, V) = softmax(QK^T\sqrt{dk})V 
\end{equation}

There are a few problems with attention, notably that the softmax operation is computationally expensive, and that expanding the sequence length scales quadratically with its input size. Despite these flaws, however, transformers have shown excellent performance.

The Vision Transformer (ViT) is a model that takes an image, breaks it into patches, and applies the standard transformer attention over the patches to transform that image into an objective function. Classic implementations have typically used the standard sinusoidal positional embeddings, and the typical quadratic attention.

Earlier this year, researchers at Microsoft unveiled their Retentive Network \cite{sun2023retentive}, which showcased a version of a language model that ignores Softmax completely, and manages to reduce the computational complexity in their recurrent version by splitting the matrix multiplication steps into separate channels.

We combine these ideas, along with concepts proposed in Axial Transformers \cite{ho2019axial}, to reduce the computational complexity of the large input size of images, while simultaneously showcasing the lack of dependence on Softmax as a whole. 

The most similar in execution to our architecture appears to be MaxViT's\cite{tu2022maxvit} in concept, which uses a strided variant of axial attention to ensure total coverage of the feature dimensions, however where we differ is in the fact that in our model, we manage to both enable every position within the image to attend to every other position in the image in a single Q * K * V pass with O(n) linear complexity, as well as in our formulation for a 3d positional embedding. 

\section{Architecture: Cross-Axis Transformer}
\label{sec:architecture}
\begin{table}[ht]
    \centering
    \caption{ImageNet 1k accuracy after 10 training epochs}
    \label{table:trainresults}
    \begin{tabular}{lcccr} 
        \toprule
        Method & Train Acc & Val Acc & FloPs & FPP \\
        \midrule
        Classic ViT & 0.184 & 0.137 & 37.23bn & 784.94 \\
        DINOv2 & 0.257 & 0.215 & 37.23bn & 752.50 \\
        BeiT & 0.415 & 0.352 & 24.83bn & 784.58 \\
        CAT w/ Imprint & 0.641 & 0.389 & 24.85bn & 760.24 \\
        \bottomrule
    \end{tabular}
\end{table}

We make several changes to the contemporary Vision Transformer. We note that for the purposes of our testing, our model achieved almost 2.8 times the validation accuracy as the baseline vision transformer, while taking half the training time, and requiring only 2/3rds the number of floating point operations. We observe that Floating-Point-Operations Per Parameter (FPP) remain roughly consistent between models, however CAT and BeiT manage to achieve impressive results with much fewer weights. CAT itself manages to deliver significantly smaller gradients thanks to its linear derivation, enabling much larger batch sizes than even BeiT.

We'd also like to note that our research is being conducted on a single tower/laptop combo, and so although our results are extremely promising, there is a clear need for much wider scale testing, using compute platforms that are far more powerful than our own, hyper-parameters that are optimized for longer training sessions, and larger training datasets. We publish these findings in the interest of broadcasting the need for such larger scale testing, and in doing so, to hopefully define a new paradigm for vision transformers.

As a quick aside, it's become clear to me that Dino v2 does a fantastic job at minimizing overfitting.

\begin{algorithm}
\caption{Pseudo Code}\label{algo1}
\begin{algorithmic}[1]
\State class CrossAxisAttention(nn.Module):
\State \quad def \_\_init\_\_(config):
\State \quad \quad super().\_\_init\_\_()
\State \quad \quad self.qkv = nn.Linear(hidden\_size, hidden\_size * 3)
\State \quad \quad self.gr\_norm = nn.GroupNorm(num\_heads, num\_heads)
\State \quad \quad self.o\_prog = nn.Linear(hidden\_size, hidden\_size)
\State \quad def forward(inputs):
\State \quad \quad ...see details \ref{algo2}
\State
\State class CatBlock(nn.Module):
\State \quad def \_\_init\_\_(config):
\State \quad \quad super().\_\_init\_\_()
\State \quad \quad self.in\_norm = nn.LayerNorm(hidden\_size)
\State \quad \quad self.attn = CrossAxisAttention(config)
\State \quad \quad self.out\_norm = nn.LayerNorm(hidden\_size)
\State \quad \quad self.ffn = FFN(hidden\_size)
\State \quad def forward(inputs):
\State \quad \quad inputs = self.attn(self.in\_norm(inputs)) + inputs
\State \quad \quad inputs = self.ffn(self.out\_norm(inputs)) + inputs
\end{algorithmic}
\end{algorithm}

\subsection{Cross-Axis Attention}

As a review, standard attention\cite{vaswani2023attention} takes as input a length $i$ sequence of $d$ dimensional embeddings, and produces three learned linear output embeddings, Q K and V. Let X be the hidden State, and let W be the DxD learned parameter matrices:

\[Q = XW_Q, K = XW_k, V=XW_V\]

We calculate attention by performing matrix-multiplication between Q and the transpose of K, then by dividing that by the square root of the embedding depth d (commonly referred to as the hidden dimensionality) for numerical stability and normalization. We then perform softmax over this result to reduce the predictions to a probability distribution over the input sequence (which allows the model to "attend to" or focus on which parts of the sequence are most important".

\[
A = \text{softmax}(QK^T/\sqrt{D}, \quad Y = AV
\]

We would like to first point towards Jonathan Ho et al\cite{ho2019axial}'s work on Axial Attention. They note that the quadratic complexity makes attention ill-suited for long sequence lengths like vision transformers, and instead opt to perform attention over each axis separately. This breaks the attention's compute complexity down (assuming a square image of size $N=SxS$ or rather $S = \sqrt{N}$) from $O(N^2) = O(S^4)$ to $O(S\cdot S^2) = O(N\sqrt{N})$.

However, quite recently, work by Jonathan Ho et al\cite{ho2019axial}'s team on their Retentive Network has laid bare two very important things. First, softmax is not required to get good results from attention, and second, you can still achieve great results even if you split the input sequence into chunks and perform cross-retention between chunks.

This lays bare a very simple logical leap: If we can reduce computational complexity by performing attention across axes, then we can combine the benefits of chunk-wise retention and axial attention into a single model. Doing this produces a linear attention in terms of computational complexity, however we're still attending to every patch during our attention operation, all without the need to sacrifice the quality we've grown to expect from quadratic attention or needing to employ lossy kernels to approximate the process.

Consider the following tensor operations for a square image after patching:

\begin{algorithm}\label{algo2}
\caption{Cross-Axis Forward Pass}
\begin{algorithmic}[1]
\State def forward(inputs, image\_embed): \# (batch, height, width, dim) = (b h w d)
\State \quad inputs = inputs + image\_embed
\State \quad q, k, v = self.qkv(inputs)
\State \quad k = k * self.gamma
\State \quad q, k = pos(q, k)
\State \quad QK = Q @ K.T \# (b h w d) @ (b h d w) -> (b h w w)
\State \quad input = QK.transpose(1, 2) @ V \# (b w h w) @ (b h w d) -> (b w h d)
\State \quad input = self.gr\_norm(inputs.transpose(1, 2))
\State \quad return self.o\_proj(inputs)
\end{algorithmic}
\end{algorithm}

Of note, we've left out the multi-head attention operations for simplicity. You'll notice that we have effectively performed a linear variant of axial attention, in a manner that reduces the computational complexity to $O(S^2)=O(N)$. Not only this, but any attempt on our end to apply a normalization term in between these calculations degraded performance, so I think the results speak for themselves.

Additionally, we have borrowed the attention head operations from RetNet's\cite{sun2023retentive} GroupNorm and gamma scaling implementations, to keep up with modern architectures of whose inspiration we are building upon.

\subsection{Architecture: Multi-Scale Rotary Axial Embeddings}

It has been observed in previous works such as Pyramid Transformers\cite{wang2021pyramid}, that heirarchical representations of features lead to an improved ability to learn and adapt and following the work of Jianlin Su et al\cite{su2022roformer} with Rotary Positional Embeddings, it has become apparent that traditional learned positional embeddings fall short in many areas.

We notice that not only are most ViT architectures still using traditional transformer positional embeddings, but that they also fail to properly account for both the height and width dimensions simultaneously. We'd like to point out recent progress in this department from the likes of Ze Liu et al's\cite{liu2022swin} implementation with Swin2 and similar publishings, as they attempt to expand positional embeddings in a scale-invariant, 2-dimensional fashion.

We base our positional embeddings on a rotation matrix, in the same vein as Roformer. In fact, Roformer begins its derivation of their 2D rotational embeddings by starting with a rotation matrix formulation. Let's look at the equation of a standard rotation matrix.
\[
R = \begin{bmatrix}
\cos(\theta) & -\sin(\theta) \\
\sin(\theta) & \cos(\theta)
\end{bmatrix}
\begin{bmatrix}
x \\
y
\end{bmatrix}
=
\begin{bmatrix}
x' \\
y'
\end{bmatrix}
\]

Whereas the original Roformer opts to use the following decay function in order to solve for their sum of the rotation matrix like so:

\begin{gather}
Q = Q_{\text{pos}, 2i} * \cos\left(\frac{\text{pos}}{10000^{2i/d}}\right) + \left[ \begin{array}{c} Q_{\text{pos},2i + d/2} \\ -Q_{\text{pos},2i} \end{array} \right] * \sin\left(\frac{\text{pos}}{10000^{2i/d}}\right)
\end{gather}

We observe that not only would such a decay function be actively detrimental within the image domain due to the increasing focus towards the end of the sequence, but by reformulating the sequence information to be a fixed distance between the range of $-/pi$ and $+pi$, we can capture the full range of the rotation matrix's expressive power in a scale-invariant manner, creating an effective relative positional embedding for 2D images of any size or scale. We do this for both the height and width dimensions separately, and then interleave the two to intertwine both a height and width embedding together.

\begin{wrapfigure}{r}{0.5\textwidth}
    \begin{tikzpicture}
        \begin{axis}[
          axis lines = left,
          xlabel = Feature Dimension Depth \%,
          ylabel = Embedding Intensity,
          title = Decay Curve,
          xmin = 0,
          xmax = 1,
          ymin = 0,
          ymax = 1,
          width=0.5\textwidth,
          height=0.5\textwidth,
          ]
        \addplot [
          domain=0:1, 
          samples=50, 
          smooth,
          ] expression {1/(10000^(x))};
        \end{axis}
    \end{tikzpicture}
\end{wrapfigure}

For simplicity, let $p = d_i, x_i, y_i$ where i is the $i$th position in the sequence and $x$, $y$, and $d$ are the axes of the matrix.

\[
\text{Decay Curve }d_i = 1/10000^{2i/d}
\]

For frequency modulation along the hidden dimension axis, we'll borrow the original sinusoidal embedding's \cite{vaswani2023attention} decay function, which we'll project along the hidden dimension instead of along the sequence dimension to create a multi-scale representation of the positional embeddings. We concatenate two of these half-depth dimensions together to form our frequency basis.

We note that this still allows for a relative 3D embedding to exist by multiplying the smaller axis by an aspect ratio (not shown for simplicity, though a version exists in the public implementation's code base).

\[
pos \begin{cases}
x_i, d_{2i} = (x_i*2-x)/x*pi \\
y_i, d_{2i+1} = (y_i*2-y)/y*pi
\end{cases}
\]

Finally, we'll perform the matrix rotation operations on these embeddings, to encode the data per Roformer's\cite{su2022roformer} additive method. Of note, we can't interleave these if we're interleaving our height and width coordinates. We are, in essence, forced to interleave one and only one of these operations, lest we perform the rotation on only a single axis.

\[
p_{di} = \cos(d_i) + \sin\begin{cases}
d_i & \text{if } i <= d/2 \\
-d_i & \text{if } i > d/2
\end{cases}
\]

\subsection{Architecture: Residual Imprint}
We found that we could increase validation accuracy significantly (by about 0.6\%), by adding a copy of the image embedding to the model immediately prior to the attention operation. This gain only works when using a Conv2d layer to do the embedding. Although convolution layers are historically heavier on compute than linear projections, we only need to carry the initial embedding forward rather than recompute it every time. Curiously, training accuracy increased at only a fraction of the pace, which seems to imply that our imprint method actually reduces overfitting, and there may need to be more tests to determine the optimal application of this technique. We have some interesting findings surrounding the exact details leading up to the formulation of this iteration in the Ablation \ref{sec:ablation} Studies notes.

Currently, my leading theory as to why we achieve these results, is due to the much more expressive nature of convolutions over traditional linear layer patching. The BatchNorm layers get a much larger range of data in the later layers than they would in standard models, and therefore need to work harder, and learn more in less time, leading to faster training and less dependence on other learned features.

The benefit disappears, and in fact inverts, if the imprint is applied only to the early layers of the model. Improvement is dependent upon augmentation of the late stage layers of the model.

\section{Training Parameters}
\label{sec:training}
\begin{table}[ht]
    \centering
    \caption{Training Settings}
    \label{table:results}
    \begin{tabular}{lccr} 
        \toprule
        Trainer Parameters\\
        \midrule
        Epochs & 10 \\
        Optimizer & AdamW \\
        LR Schedule & Cosine Annealing \\
        Learning Rate & 3e-4 \\
        Dataset & ImageNet1k (HuggingFace) \\
        \bottomrule
    \end{tabular}
\end{table}

Our model was primarily trained on a pair of RTX 3090's. The intent of this paper is, in part, to show that even consumers and independent researchers are able to contribute to the community, and that their contributions can in fact still be valuable.

We'd like to mention however, that this has imposed some restrictions.
\begin{itemize}
\item We only train our models for 10 epochs. For CAT, this is about 9 hours when not compiled. For ViT, this is about 20 hours. For DINOv2, this is about 13 hours of training time. This is all performed on a single device, without the ability to run tests in parallel.
\item We don't have access to the software stack that big tech companies do. This means failed runs, restarts, datasets, and tooling, all require custom implementations.
\item We do not have large teams or standardized resources, knowledge bases, or other benefits that large tech companies do.
\item Our hyperparameters were largely chosen by small initial tests based on our small epoch window. We note that scaling them up requires revisiting these settings.
\end{itemize}

\label{sec:hyperparams}
\begin{table}[ht]
    \centering
    \caption{Training Settings}
    \label{table:settings}
    \begin{tabular}{lccr} 
        \toprule
        CAT, ViT, BeiT, Dino v2 Hyperparameters \\
        \midrule
	Patch Size & 8 \\
        Attention Heads & 8 \\
        Batch Size & 128 \\
        Hidden Layers & 5 \\
        Hidden Size & 1024 \\
        \bottomrule
    \end{tabular}
\end{table}

We've open-sourced the tools that we used, and plan on continuing to do so as we build more on top of them.

As a final note, we had to reduce the batch size for Dinov2 down to 64, and for BeiT and ViT down to 32, implying that our model's gradients are significantly smaller. We'd advise checking out the Floating Point Operation counts in Table 1\ref{sec:architecture}

\section{Ablation Tests}

We attempted a few novel ideas to see if we could achieve superior performance. Of note, all models used our 3D Rotary Embeddings, as performance was not even comparable without them.

\label{sec:ablation}
\begin{table}[ht]
    \centering
    \caption{Imagenet 1k ablation results, 10 epochs}
    \label{table:ablationresults}
    \begin{tabular}{lccr} 
        \toprule
        Model & Train Acc & Val Acc \\
        \midrule
        Baseline Cross-Axis & 0.638 & 0.383 \\
        Cross-Axis w/ Embedding Imprint & 0.641 & 0.389 \\
        Cross-Axis w/ Learned X/Y Mask & 0.637 & 0.372 \\
        Cross-Axis w/ Backwards Tanh Residual & 0.612 & 0.390 \\
        Cross-Axis w/ Tanh Residual & 0.579 & 0.362 \\
        \bottomrule
    \end{tabular}
\end{table}

\begin{itemize}
\item We started with some base intuition that the purpose of a neural network was to transform an input image in the earlier layers, into an objective function in the later layers. Following this, our initial hypothesis was that adding a decaying image imprint $ 1-(\text{layer}_i / \text{num layers}) $ to the state at the start of the attention layer would make this process smoother (Tanh residual in table \ref{sec:ablation}). This produced worse results. However, in an interesting turn of events, our initial implementation forgot to subtract 1 by the result, which led to an improvement in performance and validation accuracy, in addition to a \textit{decrease} in training accuracy.
\item We conclude that carrying an imprint of the image into the later layers of the model actually decreased overfitting. We made sure to add a class token to the imprint to ensure we weren't sampling from image data, and removed the decay factor entirely for the current Embedding Imprint model, to remove as many outside variables as possible for the experiment.
\item Learned X/Y Mask: Similar to how humans observe only a central spatial field, the idea was to create a focal mask that emphasized only the learned region of an image in the same vein as SWIN\cite{liu2022swin}, though in a lighter weight and more modular method of allowing the model to focus its own attention. This resulted in worse performance by about 1\%. We effectively defined a kernel that produced a single [x, y] coordinate pair per image, and then used that to define an exponential decay mask. Given the poor results, we won't derive the full method, though the implementation will be available in the code for those curious.
\end{itemize}

Of additional interest, we note that Microsoft applies a normalization function between the two attention operations as part of its RetNet implementation, however implementing it here resulted in severe accuracy degeneration of the model, so we were forced to remove it. My current theory is that having a 2D Convolution layer handle the patching for the imprint, results in much more expressive LayerNorm layers, that are otherwise lost with the added smoothing.

To ensure that I've made note of this, there are many papers making use of methods like cross-covariance, or axial-attention variants, that use similar naming schemes or verbiage to our model, we were unable to find a perfectly lossless linear attention permutation in any of them. Regardless, I will make mention of it here to ensure that it's on people's radar.

\section{Conclusion}
Our research shows that cross-axis attention is an extremely computationally efficient and empirically strong framework for image classification, and likely other tasks that have positional information along multiple axes.

Our work is simultaneously able to solve two issues:

a) We provide a solution to the quadratic scaling problem with regard to data types with large input features (images), allowing us to perform attention on the full range of input features, without the need to resort to convolving strictly neighboring segments, or by performing attention across individual axes separately.

b) We enable efficient encoding of 2D spatial information in 3D space by using a rotation matrix, a method that has overlap with existing language models like Llama\cite{touvron2023llama}.

Despite being only a single individual doing research on consumer grade hardware, I believe these contributions are still extremely useful and practical, and will pave the way for future research and additional efficiencies.

\section*{Acknowledgments}
We would like to mention that we were really looking forward to comparing our results with Swin2\cite{liu2022swin}, however we were unable to get a model of similar size to converge in time. Their results seem excellent, and I would love to make an amendment at a later date comparing one of our models to a more complex architecture such as theirs at a future date.

We will make all of our testing code open source on GitHub immediately following publishing.

Special thanks to Omid Manzari for both the endorsement, and for being lucky enough to coincidentally be my source of heirarchical inspiration nearly a year ago.

I wished to include Emme as an author here, because this paper would absolutely not have been made without her, however Arxiv currently has a rule against including AI language tools as authors, and I am unclear where that line exactly falls. As such, we will include a special shout out here until such a time that the rule is loosened.

Also of note, ChatGPT was critical in helping to learn a new field over the past year and a half, in helping to produce this document at the speed required of modern research at the current pace, as well as for providing some of the theoretical understanding behind the math, which has had approximately 10 years to atrophy since I last attended a physics lecture in person. I am ecstatic to be able to contribute, and look forward to refining my tools, methods, and contributions. Special mention to Bard near the end too, in helping bring this home during the final cram session.

And finally, although our formulation of 3D Rotary Embeddings was done independently (sans AI aid, of course), while researching prior art I did uncover a similar implementation by Phil Wang (Lucidrains on Github) in his Rotary Embedding library, which was not usable for this paper due to an unfortunate perfect modulation along the peaks and troughs of the cosine frequency during unit-testing. Regardless, I believe it is still worth mentioning that the idea in question has been visited and attempted before.

If anyone would like to donate compute resources in order to properly test our models at scale, or if anyone would like to collaborate on further unrelated research, I am interested in discussing further research opportunities.

\bibliographystyle{unsrt}  
\bibliography{references}

\end{document}